\documentclass[runningheads]{llncs}
\usepackage{amsmath}
\usepackage{amssymb}
\usepackage{booktabs}
\usepackage{hyperref}
\usepackage[T1]{fontenc}
\usepackage{graphicx,verbatim}
\let\oldthebibliography\thebibliography
\renewcommand{\thebibliography}[1]{%
  \oldthebibliography{#1}%
  \setlength{\itemsep}{-2pt}%
  \setlength{\parsep}{0pt}%
}
\begin{document}
\title{Medical Image Spatial Grounding with Semantic Sampling\thanks{Preprint. Accepted at MICCAI 2026. This is the authors' version of the work.}}
%
\author{Andrew Seohwan Yu$^{1,2,\dagger}$ \and
Mohsen Hariri$^{1,\dagger}$ \and
Kunio Nakamura\inst{2} \and
Mingrui Yang\inst{2} \and
Xiaojuan Li\inst{1,2} \and
Vipin Chaudhary\inst{1}}
\index{Yu, Andrew Seohwan}
\index{Hariri, Mohsen}
\index{Nakamura, Kunio}
\index{Yang, Mingrui}
\index{Li, Xiaojuan}
\index{Chaudhary, Vipin}
\authorrunning{A. Yu et al.}
%
\institute{Case Western Reserve University, Cleveland OH 44106, USA \and
Cleveland Clinic, Cleveland OH 44106, USA\\[3pt]
$^{\dagger}$\,These authors contributed equally.\quad
Corresponding author: Vipin Chaudhary (\email{vipin@case.edu}).}

\maketitle              
\begin{abstract}
Vision language models (VLMs) have shown significant promise in visual grounding for images as well as videos. In medical imaging research, VLMs represent a bridge between object detection and segmentation, and report understanding and generation. However, spatial grounding of anatomical structures in the three-dimensional space of medical images poses many unique challenges. In this study, we examine image modalities, slice directions, and coordinate systems as differentiating factors for vision components of VLMs, and the use of anatomical, directional, and relational terminology as factors for the language components. We then demonstrate that visual and textual prompting systems such as labels, bounding boxes, and mask overlays have varying effects on the spatial grounding ability of VLMs. To enable measurement and reproducibility, we introduce \textbf{MIS-Ground}, a benchmark that comprehensively tests a VLM for vulnerabilities against specific modes of \textbf{M}edical \textbf{I}mage \textbf{S}patial \textbf{Ground}ing. We release MIS-Ground to the public at \href{https://github.com/asy51/mis-ground}{\texttt{github.com/asy51/mis-ground}}. In addition, we present \textbf{MIS-SemSam}, a low-cost, inference-time, and model-agnostic optimization of VLMs that improves their spatial grounding ability with the use of \textbf{Sem}antic \textbf{Sam}pling. We find that MIS-SemSam improves the accuracy of Qwen3-VL-32B on MIS-Ground by 13.06\%.


\end{abstract}

\section{Introduction}\label{sec:intro}
Merging the embedding spaces of vision and language models through contrastive learning \cite{clip} was a revelation that gave rise to the modern vision language model (VLM). Video VLMs were the logical next step, extending these ideas to spatiotemporal data where models must align sequences of frames with natural language. Despite the sheer scale of information in videos relative to images, substantial progress has been made in both video understanding and multimodal reasoning \cite{videollava}.

Medical imaging VLMs have been a different matter. As narrow vision models continue to mature in the medical field, medical VLMs are in their early stages of development. Because the available data is limited, models, benchmarks, and training objectives tend to revolve around report generation \cite{ctrate,mimiccxr} and visual question answering \cite{pathvqa,slake,vqarad,medgemma}, especially disease identification. Though the reported results can be impressive, it is difficult to gauge the practicality of these VLMs in a setting as high-stakes as the medical field. Sambara et al. \cite{3dreasonknee} showed that despite using the correct terminology and making references to anatomical components and lesions, VLMs failed to display reasoning capabilities that were spatially grounded in objects that were actually present in the image in question.

There has been recent work to introduce rigor in spatial grounding to medical VLM training and benchmarking, demonstrating that general-purpose 2D VLMs fail at even simple visual grounding of medical images, due in large part to language-only pretraining that contains ground truths about human anatomy, which may differ from the radiological image in question \cite{mirp}. Methods for linking anatomical components to language using contrastive alignment in 2D images have also been introduced \cite{anatomix}.

\begin{figure}
\includegraphics[width=\textwidth]{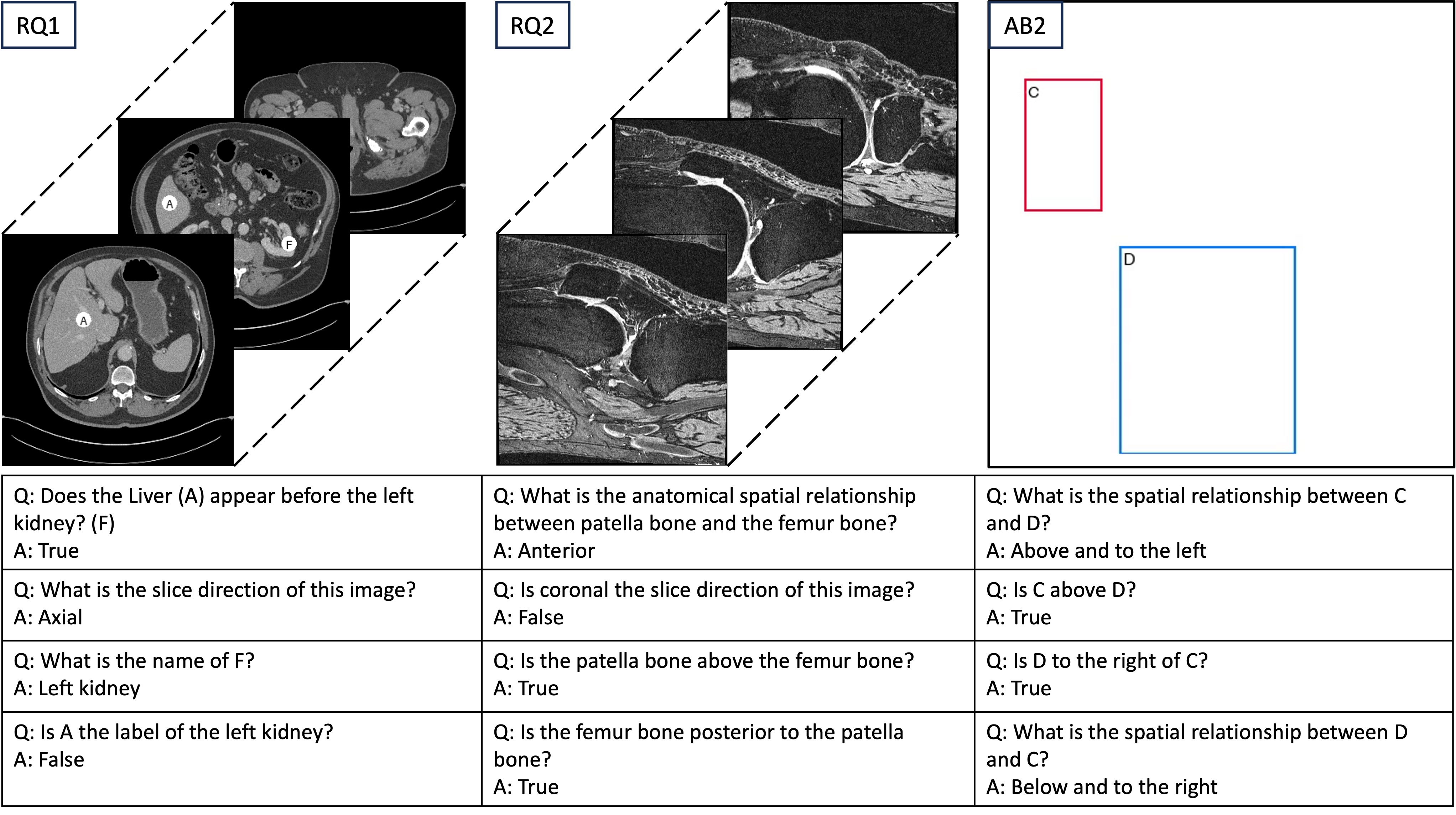}
\caption{Sample question-answer pairs from MIS-Ground. For each 2D or 3D vision input (top), multiple questions are generated (bottom). Video VLMs display the ability to discern anatomical structure relationships across slices (RQ1); they prefer anatomical direction terms like anterior (RQ2); and they are capable of abstract spatial reasoning (AB2). Note: the CT image is shown in axial RAS standard viewing orientation, whereas the MRI is shown in sagittal RAS storage mode.}
\label{examples}
\end{figure}

To push the current understanding of 3D VLM spatial grounding capabilities, this study proposes \textbf{M}edical \textbf{I}mage \textbf{S}patial \textbf{Ground}ing (\textbf{MIS-Ground}), a benchmark designed to identify the extent to which video VLMs are able to identify, localize, and differentiate anatomical structures in 3D medical images. MIS-Ground covers a wide variety of language and vision inputs that are likely to occur in medical imaging scenarios. These include different text and visual prompts to align the object of interest to text, multiple views of the same 3D volume, as well as various combinations of question classes and target answers (Figure \ref{examples}). Because a 3D volume can be serialized as an ordered stack of slices, video-native VLMs offer a practical interface for it, mirroring how video segmentation models have been repurposed for 3D MRI by treating slices as video frames \cite{yu2024sam2knee}; MIS-Ground is nonetheless not video-specific, and we also evaluate 3D-enabled medical VLMs (Section \ref{sub:eval}). To our knowledge, MIS-Ground is the first benchmark to probe 3D anatomical spatial grounding under a controlled, factorial design that jointly varies modality, slice direction, coordinate convention, prompt type, and terminology, whereas prior medical-grounding studies report aggregate failures \cite{3dreasonknee,mirp} without isolating these factors. This study systematically evaluated five key dimensions of spatial grounding, organized as three research questions (RQ) and two ablations (AB): (RQ1) their capacity for 3D understanding across slice directions; (RQ2) their preference for anatomical over colloquial direction terms; (RQ3) the conditional impact of visual prompts based on orientation and terminology; (AB1) their reliance on pretrained anatomical knowledge priors; and (AB2) their ability to perform abstract spatial reasoning in the absence of medical images.

This study also proposes \textbf{M}edical \textbf{I}mage \textbf{S}patial \textbf{Sem}antic \textbf{Sam}pling (\textbf{MIS-SemSam}), which enhances the reasoning capabilities of the models by employing a sampling strategy based on groups of semantically similar tokens instead of individual ones. This study tested open-weights models Qwen \cite{qwen,qwen3,qwen25vl,qwen2vl,qwenvl}, MedGemma \cite{medgemma}, Molmo \cite{molmo}, and paid model Gemini \cite{gemini25,gemini3flash} on MIS-Ground; M3D \cite{m3d} and Med3DVLM \cite{med3dvlm}, two leading 3D medical VLMs, were also tested. With an overall accuracy of 66.5\%, Qwen3-VL-32B with MIS-SemSam demonstrated a strong ability in 3D medical image spatial grounding, despite being trained on videos of natural scenes.

\section{Methods}
MIS-Ground was designed to systematically test the spatial grounding capabilities of VLMs against a variety of scenarios in medical imaging. This requires a careful curation of the dataset and data pipeline to query the VLMs effectively and efficiently. These integrations are included as part of MIS-Ground. VLMs with and without MIS-SemSam are treated as separate models and thus can be tested on MIS-Ground individually.

\subsection{Dataset Preparation}
Knee MRI from the Osteoarthritis Initiative (OAI) \cite{oai} and torso CT from TotalSegmentator (CT dataset) \cite{wasserthal2023totalsegmentator} were used in this study. With access to the MRI dataset under the appropriate data use agreement, 209 double echo steady state (DESS) MR images of the knee joint were collected. For each of these scans, 3D bounding boxes for 10 anatomical components in the knee were gathered from 3DReasonKnee-Bench \cite{3dreasonknee}, an open license (CC BY 4.0) benchmark dataset. These components included bone, cartilage, ligaments, and other regions in the knee joint that are associated with osteoarthritis. For the CT dataset, 1228 full-dose axial scans of the torso were collected partly under an open license (CC BY 4.0) and a non-commercial license from TotalSegmentator \cite{wasserthal2023totalsegmentator}. Under the same licenses, up to 117 matching segmentation masks of anatomical components were collected for each scan. Prioritizing visibility and ease of prompting, 67 of the larger and more commonly named components were selected for this study, and scans with fewer than 200 or greater than 450 slices were omitted. For preprocessing, MRI and CT datasets underwent percentile-based and soft-tissue windowing, respectively. Multi-planar reconstruction with trilinear interpolation was used to render the scans in alternative slice directions. Finally, the scans were placed in one of two orientations: RAS (right-anterior-superior) coordinate system storage (where \([0,0,0]\) is the RAS-most point) and standard viewing orientations.

\subsection{Benchmark Curation}
MIS-Ground consists primarily of questions evaluating the relative positions of anatomical structures. Questions were defined by four parameters: (1) Visual prompts were derived from ground-truth annotations, including masks, bounding boxes, and centroids (points) for the CT dataset, and bounding boxes for the MRI dataset. Prompts were uniquely colored and/or labeled with a letter A-F. (2) Text prompts included names of anatomical structures and/or the colors or letters of the visual labels. (3) Target type determined the target answers of the generated queries: structure names, labels, or spatial relationships, classified as either anatomical (e.g. superior) or colloquial (above). (4) Question type formatted the queries as open-ended, closed (True), or closed-inverted (False). Ablations included omitting visual inputs entirely (text-only) or omitting the medical image itself (visual prompts on a white background). To ensure broad coverage, pairs of anatomical structures were randomly selected for each parameter combination, yielding an average of 683 and 2,757 questions per MRI and CT scan, respectively.

\subsection{VLM Evaluation}\label{sub:eval}
A variety of video-native open-weights VLMs were benchmarked on MIS-Ground, including Alibaba's Qwen2.5-VL-Instruct (3B, 7B, 32B, 72B) \cite{qwen25vl} and Qwen3-VL-Instruct (2B, 4B, 8B, 32B) \cite{qwen}; and AllenAI's Molmo 2 (4B, 8B) \cite{molmo}.  In addition, 3D-enabled medical VLMs were tested, including Google's MedGemma (4B, 27B) and MedGemma 1.5 (4B) (based on Gemma 3) \cite{medgemma}; M3D (based on Llama 2 7B) \cite{m3d}; and Med3DVLM (based on Qwen2.5 7B) \cite{med3dvlm}. For individual model performances, see Figure \ref{graph}. Additionally, closed-source models, Google's Gemini models (2.5 Flash, 2.5 Flash Lite, 3 Flash) were benchmarked as well. To ensure fairness in benchmark comparisons and to promote reproducibility, inference configurations were tuned for the same sampling configuration: reasoning with max new tokens=8,912 and temperature=0.5. To report performance robustly under limited trials, we report the Bayesian credible interval \cite{hariri2026dont}. Due to model design, inputs to M3D and Med3DVLM were resized to \([32, 256, 256]\) and \([128,256,256]\) respectively with trilinear interpolation, and may have resulted in reduced performance.

\subsection{Semantic Sampling}
As discussed in Section \ref{sec:intro}, many failures in medical image spatial grounding are not due to a complete lack of visual recognition, but rather to \textbf{language-side brittleness}: small decoding perturbations can flip directional terms (e.g., \textit{superior} vs.\ \textit{inferior}), select awkward morphological variants, or drift into semantically adjacent but incorrect anatomical phrasing. This effect becomes more pronounced under long-form reasoning (chain of thought), where a single locally plausible token can derail the subsequent trajectory. To mitigate this issue at inference time without additional training, we introduce \textbf{MIS-SemSam}, an adaptation of Semantic Sampling to medical VLM spatial grounding.

In Semantic Sampling, instead of selecting the next token purely by its own probability, Semantic Sampling prefers tokens that are supported by probability mass in their local semantic neighborhood in token-embedding space. Intuitively, if the model is unsure among several semantically similar realizations of the same concept (common in directional and anatomical terminology), aggregating neighborhood mass stabilizes decoding toward semantically coherent regions of the vocabulary.

\textbf{Offline semantic neighborhoods.}
Let the language component of a VLM have an input embedding matrix $E\in\mathbb{R}^{V_{\text{emb}}\times d}$ and a tokenizer vocabulary $V_{\text{tok}}$. Following Semantic Sampling decoding, we partition token ids into \emph{content tokens} $C$ (ordinary text) and \emph{non-content tokens} $U$ (special, added or control tokens and any reserved embedding rows not produced by the tokenizer). This distinction is particularly important for VLMs because modality and chat-template markers (e.g., image delimiters, role tokens) behave as control symbols and can form embedding space ``hubs'' that contaminate nearest-neighbor structure.

We therefore construct neighborhoods \emph{only over content tokens}. Let $E_C = E[C]$ denote the content-token embeddings. We row-normalize
\begin{equation}
\hat{e}_i \;=\; \frac{E_C[i]}{\lVert E_C[i]\rVert_2 + \epsilon}, \quad i\in C,
\end{equation}
so that cosine similarity is $\cos(i,j)=\hat{e}_i^\top\hat{e}_j$. For each $i\in C$, we precompute its $K$ nearest neighbors in $C$ under cosine similarity (including $i$ itself) and store two lookup tables:
\begin{equation}
S_{\text{tid}}\in\mathbb{N}^{|C|\times K},\quad S_{\text{val}}\in\mathbb{R}^{|C|\times K},
\end{equation}
where $S_{\text{tid}}[i,k]$ is the $k$-th neighbor id and $S_{\text{val}}[i,k]$ its cosine similarity. These tables are computed once per embedding space and reused across all MIS-Ground queries; moreover, when multiple checkpoints share the same tokenizer indexing, the same neighbor \emph{ids} can be reused as in Semantic Sampling.

\textbf{Inference-time semantic rescoring.}
At decoding step $t$, the VLM outputs logits $\ell_t\in\mathbb{R}^{V_{\text{emb}}}$ over the next text token, which induce (temperature-scaled) probabilities
\begin{equation}
p_t(v)=\frac{\exp(\ell_t(v)/T)}{\sum_{v'} \exp(\ell_t(v')/T)}.
\end{equation}
We first apply any standard truncation filter $\textsc{Filter}(\cdot)$ (e.g., top-$M$, top-$p$) to obtain a candidate set $I_t\subseteq V_{\text{emb}}$.
Because semantic neighborhoods are defined only for content tokens, if $I_t\cap U\neq\emptyset$, we skip MIS-SemSam for this step and defer to the model's default decoding rule (greedy for $T{=}0$, or the configured sampler for $T{>}0$). Otherwise, $I_t\subseteq C$ and we compute a semantic score for each candidate $c\in I_t$ by aggregating probability mass over its neighborhood:
\begin{equation}
\text{Score}_t(c) \;=\; \sum_{k\in\mathcal{K}(c)} w_{c,k}\, p_t\!\big(S_{\text{tid}}[c,k]\big),
\qquad
w_{c,k}=\max\!\big(0,\, S_{\text{val}}[c,k]\big),
\end{equation}
where $\mathcal{K}(c)$ denotes the kept neighbor indices (either the top-$K'$ neighbors, $K'\le K$, or a similarity-threshold prefix), and negative cosine values are clamped to zero to prevent cancellation.

Finally, the next token is chosen from $I_t$ using either:
(i) deterministic selection $x_{t+1}=\arg\max_{c\in I_t}\text{Score}_t(c)$ (used to maximize reproducibility), or
(ii) stochastic selection by sampling from a softmax over $\text{Score}_t$ (used in chain of thought settings to preserve controlled exploration).
MIS-SemSam requires no additional forward passes and adds only $O(|I_t|\cdot|\mathcal{K}(c)|)$ table lookups per token.

MIS-SemSam is applied as a drop-in replacement for the final ``pick next token'' step in generation, leaving the vision encoder, prompting strategy, and truncation filter unchanged. In our experiments, we treat a model with MIS-SemSam enabled as a separate inference configuration (Section \ref{sub:eval}), enabling apples-to-apples comparisons between default decoding and semantic-neighborhood-supported decoding on identical MIS-Ground questions. The reported improvement is therefore a paired comparison by construction: the model, prompts, images, questions, and inference settings are held fixed, and only the final token-selection rule changes. We use \emph{model-agnostic} in the algorithmic sense---MIS-SemSam applies to any VLM that exposes next-token logits and input embeddings, and requires neither training nor additional forward passes---rather than as a claim of broad empirical validation; our strongest empirical evidence is on the Qwen3-VL family, while MIS-Ground itself spans many model families.

\section{Results}
\label{results}
A total of 1160 3D volumes, 2320 2D slices, and 33864 questions were generated across both CT and MRI subsets, which contained 67 and 10 labeled anatomical components per sample respectively. In an effort to balance the dataset for each anatomical component, the CT subset was sampled more and accounted for 84\% of the entire dataset. Consequently, the aggregate score is CT-weighted, and results should be read at the benchmark level rather than as uniform cross-modality or clinical-readiness claims. Sample question-answer pairs are shown in Figure \ref{examples}. Model response was instructed to be placed at the end in special tags, after all the reasoning tokens. If the tags were missing, the question was omitted from analysis; this omission reflects instruction-following rather than spatial correctness, and predominantly affected the smallest models and the 3D medical VLMs M3D and Med3DVLM, whose outputs were frequently malformed.

\begin{figure}
\includegraphics[width=\textwidth]{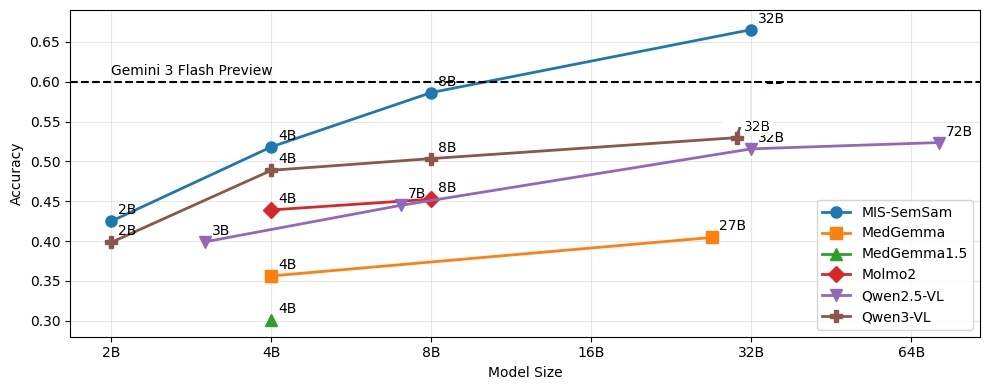}
\caption{Overall accuracy on the MIS-Ground benchmark, by model family and size. Note that Med3DVLM (0.0\%) and M3D (15.9\%) were tested but omitted from this graph. The MIS-SemSam family is derived from the Qwen3-VL family.}
\label{graph}
\end{figure}

Model version and size was strongly correlated to overall performance, with Qwen3-VL-32B outperforming all other open-weights models by a comfortable margin. The overview of these results are found in Figure \ref{graph}. M3D and Med3DVLM both failed to give valid responses, often giving long-winded explanations for unrelated concepts, often in non-English characters. These failures likely reflect aggressive input resizing and rigid response formatting (Section \ref{sub:eval}) rather than an absence of spatial capability, and should not be read as a general verdict on 3D medical VLMs. By contrast, the closed-source Gemini models served as a strong reference point, underscoring that MIS-Ground is useful beyond our own method. Medgemma did manage to follow the prompt for the most part, though it still gave poor responses. On the other hand, Qwen and Molmo models had much more success, though the very small models (2B, 3B and 4B) still struggled to give valid responses. Large Qwen2.5 models, especially 72B, underperformed relative to its size, which confirmed Qwen's iterative improvements between versions. Otherwise, no strongly discernible patterns emerged regarding model class and specific performances. As the best performing open-weights model, MIS-SemSam (Qwen3-VL-32B with semantic sampling) was used to evaluate specific spatial grounding capabilities across the following research questions and ablations.
\newline
\newline
\noindent\textbf{(RQ1) 3D Understanding}: MIS-SemSam effectively processed 3D medical images, accurately predicting slice direction in 77.6\% of cases. The model successfully determined spatial relationships across slices (68.0\% accuracy) nearly as well as in-plane relationships (71.7\%). Overall performance on 3D images (56.9\%) remained competitive with 2D images (58.8\%).
\newline
\newline
\noindent\textbf{(RQ2) Direction Terms}: The model demonstrated a strong preference for anatomical directional terms over colloquial terms. In standard viewing orientations, anatomical terms outperformed colloquial terms (69.4\% vs. 57.8\%). However, the model shifted its preference to colloquial terms (59.85\% vs. 50.2\%) in RAS storage mode, where standard medical priors conflict with the visual presentation.
\newline
\newline
\noindent\textbf{(RQ3) Visual Prompts}: The impact of visual prompts varied depending on orientation and terminology. In standard RAS viewing orientation, adding points, bounding boxes, or masks yielded only modest improvements (+2.97\%, +2.33\%, and +2.52\% respectively) over the 57.6\% baseline. For RAS storage mode utilizing anatomical terms, visual prompts severely degraded a strong 75.3\% baseline performance (-13.33\% for points, -9.22\% for boxes) due to conflicting information. Conversely, when forced to use colloquial terms, visual prompts significantly aided the model, improving a weak 45.1\% baseline by up to +8.23\%.
\newline
\newline
\noindent\textbf{(AB1) Anatomy Prior}: The model relied heavily on pretrained anatomical knowledge. In text-only ablations (no image), it achieved 69.3\% accuracy on anatomical relationships (compared to 74.0\% with the image). This prior did not extend to colloquial terms, where text-only accuracy dropped to 40.4\%.
\newline
\newline
\noindent\textbf{(AB2) Abstract Reasoning}: The model demonstrated capable abstract spatial reasoning. When tested on visual prompts placed against a blank background without medical scans, the model still accurately determined spatial relationships with 64.2\% (points) and 60.4\% (boxes) accuracy.

\section{Conclusions}
With MIS-Ground, VLMs demonstrated measurable capabilities in spatial grounding of anatomical structures in medical imaging. This is in contrast to previous studies, which reported apparent random guessing and failure to follow instructions \cite{3dreasonknee,mirp}. This is due to careful considerations made in the construction of MIS-Ground, which prioritized the balance and diversity of image modality, visual prompts, text prompts, target answer types, question types, image orientation, slice direction, 2D and 3D, terminology use, and ablations. MIS-SemSam also represents an exploration into the optimization of the VLMs in medical image spatial grounding: an inference-time method with negligible costs and significant performance improvements.

More improvements and applications of VLMs lie ahead. For one, fine-tuning is a crucial task for a domain as highly specialized as medical imaging. Unfortunately, many current such paradigms seem to overfit the task of existing medical visual question answering benchmarks and fail to translate to general capabilities like spatial grounding. New developments in such benchmarks and training datasets are necessary for this field to continue to grow. Also, spatial grounding of anatomical structures is only a partial coverage of what is needed in clinical practice, such as report generation. There is still a lack of alignment between the model capabilities and practical use. To help the community build toward these goals, we release our benchmark MIS-Ground to the public at \href{https://github.com/asy51/mis-ground}{\texttt{github.com/asy51/mis-ground}}.

\begin{credits}
\subsubsection{\ackname} This research was supported in part by NSF awards 2117439 and 2320952.

\subsubsection{\discintname} The authors declare they have no competing interests.
\end{credits}

\bibliographystyle{splncs04unsrt}
\bibliography{main}
%




\end{document}